\documentclass[runningheads]{llncs}
\usepackage[T1]{fontenc}
\usepackage{graphicx}
\usepackage{caption}
\usepackage{subcaption}
\usepackage{amssymb}
\usepackage{multirow}
\usepackage{tabularx}
\usepackage{hyperref}
\begin{document}
\title{The Weighting Game: Evaluating Quality of Explainability Methods} 
%
%

\author{Lassi Raatikainen \and
Esa Rahtu}
\institute{Tampere University, Tampere, Finland.
\email{raatikainenlassi@gmail.com esa.rahtu@tuni.fi}}
%
\maketitle              
%

\begin{center}
\small\textbf{Preprint notice.} This preprint has not undergone peer review (when applicable) or any post-submission improvements or corrections.
The Version of Record of this contribution is published in \emph{Image Analysis. SCIA 2025. Lecture Notes in Computer Science (LNCS), vol 15726} and is available online at \url{https://doi.org/10.1007/978-3-031-95918-9_23}.
\end{center}

\begin{abstract}

The objective of this paper is to assess the quality of explanation heatmaps for image classification tasks. To assess the quality of explainability methods, we approach the task through the lens of \emph{accuracy} and \emph{stability}.

In this work, we make the following contributions. Firstly, we introduce the \emph{Weighting Game}, which measures how much of a class-guided explanation is contained within the correct class' segmentation mask. Secondly, we introduce a metric for explanation stability, using zooming/panning transformations to measure differences between saliency maps with similar contents.

Quantitative experiments are produced, using these new metrics, to evaluate the quality of explanations provided by commonly used CAM methods. The quality of explanations is also contrasted between different model architectures, with findings highlighting the need to consider model architecture when choosing an explainability method.

\end{abstract}
\section{Introduction}

\begin{figure}[!t]
\centering
\begin{subfigure}[t]{.3\linewidth}
    \centering
    \includegraphics[width=\textwidth]{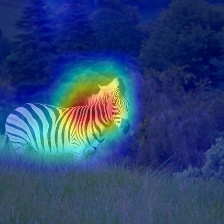}
    \caption{Class-guided explanation with class 'zebra'.}\label{fig:zebra1}
\end{subfigure}
    \hfill
\begin{subfigure}[t]{.3\linewidth}
    \centering
    \includegraphics[width=\textwidth]{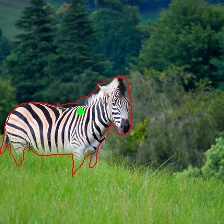}
    \caption{Pointing Game with $acc=100\%$.}\label{fig:zebra2}
\end{subfigure}
   \hfill
\begin{subfigure}[t]{.3\linewidth}
    \centering
    \includegraphics[width=\textwidth]{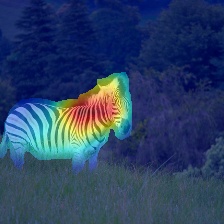}
    \caption{Weighting Game with $acc=61.6\%$.}\label{fig:zebra3}
\end{subfigure}

\bigskip
\begin{subfigure}[t]{.3\linewidth}
    \centering
    \includegraphics[width=\textwidth]{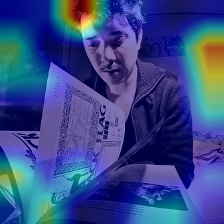}
    \caption{Class-guided explanation with class 'person'.}\label{fig:person1}
\end{subfigure}
    \hfill
\begin{subfigure}[t]{.3\linewidth}
    \centering
    \includegraphics[width=\textwidth]{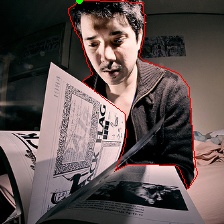}
    \caption{Pointing Game with $acc=100\%$.}\label{fig:person2}
\end{subfigure}
   \hfill
\begin{subfigure}[t]{.3\linewidth}
    \centering
    \includegraphics[width=\textwidth]{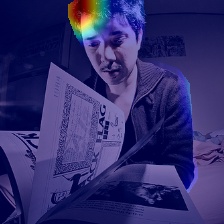}
    \caption{Weighting Game with $acc=14.7\%$.}\label{fig:person3}
\end{subfigure}
\caption{Differences of Weighting Game to Pointing Game. Top explanation is distributed better over the correct class, while bottom explanation is spread out more over the image. With Pointing Game, the accuracy results are the same for both explanations, as the highest magnitude pixel (visualized with green circle) is within the class segmentation mask. Weighting Game aims to better explain, how accurate explanations are by calculating the proportion of the magnitude of the saliency contained within the correct class mask. Class-guided explanations were calculated using ViT-B/32 \cite{ViT} and Layer-CAM \cite{LayerCam}}
\label{fig:pointing-game-hits}
\end{figure}

The rise of deep learning in the field of computer vision has seen models become increasingly powerful, however, this has been with the cost of interpretability. Due to this, the field of explainable AI has recently seen a rise into prominence. Several explainability methods have been proposed, which aim to explain what parts of an input image are most relevant to the given output of a model \cite{GuidedBackProp,GradCAM,GradCAMPLUS,AblationCam,LayerCam,XGradCam}, the output of an explainability method is usually called a \emph{saliency map}. These methods can be used to evaluate the quality of models' predictions, and especially can be useful when trying to understand incorrect predictions.

Previous works in evaluating saliency maps has found that many explainability methods are independent of the model parameters and labeling of training data, thus rendering them to be simple edge detectors \cite{SanityChecks}. While it is necessary for an explainability method to be independent of such factors, it does not imply a method is accurate in its explanations, as the sanity checks do not measure how explanations align with input images. Our goal in this paper is to build a framework for measuring the accuracy and stability of explanations. By accuracy, we refer to the ability of a class-guided explanation to correctly concentrate on the correct class' objects, while with stability we refer to the consistency of explanations when minor transformations are applied to the image.

For assessment of explainability method accuracy, mainly two types of metrics have been presented. Pointing Game \cite{ExcitationBP} measures the accuracy at which a saliency map's highest magnitude pixel is contained within the correct class's segmentation map, thus recording individual instances as either hits or misses. The problem with this method is that the binary nature of the metric does not allow for measuring the quality of hits and misses. An example of the limitations of Pointing Game is shown in Figure. \ref{fig:pointing-game-hits}, where both explanations are counted as equally good, while the one found in Figure \ref{fig:zebra1} seems more concentrated on the correct class compared to the one shown in Figure. \ref{fig:person1}.

The second types of methods for measuring explanation accuracy usually operate by removing patches or points from an image, to record how the output changes based on removal of different points \cite{LRP,Xrai,RISE}. However, removal based methods tend to introduce unnatural patches into images, which can cause undesired effects into the results, the methods also have the tendency of being more computationally intensive.

\begin{figure}[!t]
\centering
\begin{subfigure}[t]{.3\linewidth}
    \centering
    \includegraphics[width=\textwidth]{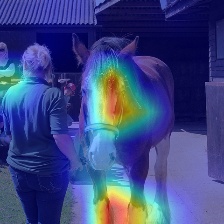}
    \caption{Class-guided explanation with class 'horse'.}\label{fig:horse1}
\end{subfigure}
    \hfill
\begin{subfigure}[t]{.3\linewidth}
    \centering
    \includegraphics[width=\textwidth]{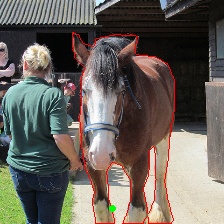}
    \caption{Pointing Game with $acc=0\%$.}\label{fig:horse2}
\end{subfigure}
   \hfill
\begin{subfigure}[t]{.3\linewidth}
    \centering
    \includegraphics[width=\textwidth]{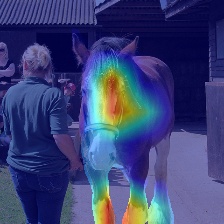}
    \caption{Weighting Game with $acc=64.6\%$.}\label{fig:horse3}
\end{subfigure}

\bigskip
\begin{subfigure}[t]{.3\linewidth}
    \centering
    \includegraphics[width=\textwidth]{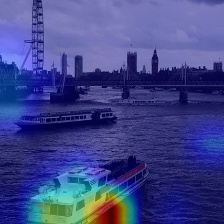}
    \caption{Class-guided explanation with class 'boat'.}\label{fig:boat1}
\end{subfigure}
    \hfill
\begin{subfigure}[t]{.3\linewidth}
    \centering
    \includegraphics[width=\textwidth]{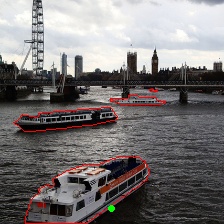}
    \caption{Pointing Game with $acc=0\%$.}\label{fig:boat2}
\end{subfigure}
   \hfill
\begin{subfigure}[t]{.3\linewidth}
    \centering
    \includegraphics[width=\textwidth]{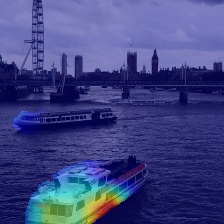}
    \caption{Weighting Game with $acc=52.6\%$.}\label{fig:boat3}
\end{subfigure}
\caption{Two examples of explanations and their accuracies. In both class-guided explanations, Pointing Game counts the explanations as misses, due to barely missing the segmentation mask. Weighting Game is able to correctly identify both explanations as quite accurate. Class-guided explanations were calculated using ViT-B/32 \cite{ViT} and Layer-CAM \cite{LayerCam}}
\label{fig:pointing-game-misses}
\end{figure}

In this paper, we extend the idea of Pointing Game and introduce the Weighting Game, which measures how much a saliency map's \emph{mass} or \emph{magnitude} is contained within the correct class's segmentation map. The measurement of the distribution of the explanation map allows for more flexibility in assessing the quality of individual explanations, as it is not binary in its nature, such as the Pointing Game. Pointing Game can slightly miss a target and define an explanation as a miss, while Weighting Game is able to see the whole picture and give a better representation of the accuracy of the explanation, such as in Figure \ref{fig:pointing-game-misses}.

For assessing the stability of an explainability method, we introduce two separate methods. Firstly, we create 3D effect videos from an image, and observe the correlation between saliency maps from consecutive frames. With sufficiently small changes between frames, this grants us the ability to assess how minor changes can affect the resulting saliency maps. Secondly, we create a pre-defined zoom and pan transformation on an image, and process saliency maps for the original and transformed images. Then, we measure the correlations between the saliency map for the transformed image and a correspondingly transformed saliency map for the original image. These methods provide a way to assess how consistent explainability methods are, thus giving insights on how well they can function in real-world use cases

The main contributions of this work are: (\textbf{1}) The Weighting Game, a novel metric for measuring the accuracy of an explainability method; (\textbf{2}) two novel metrics for assessing the stability of an explainability method; and (\textbf{3}) comparison of several prominent CAM methods using proposed metrics with commonly used image classification architectures.

\section{Related Work}

\subsubsection{Backpropagation Explainability Methods}

First methods to visualize network attribution were based on gradients of a model found via backpropagation. These methods include DeConvNet \cite{DeConvNet}, Network Saliency \cite{SimonyanSaliency}, and Guided Backpropagation \cite{GuidedBackProp}. The methods have slight difference, however, all are based in the idea of visualizing gradients of a loss function from a forward propagated input image. Later introduced backpropagation methods include SmoothGrad \cite{SmoothGrad}, Integrated Gradients \cite{IntegratedGradients}, and Expected Gradients \cite{ExpectedGradients}. 

\subsubsection{Class Activation Explainability Methods}

The use of class activation maps was first introduced with CAM as a way to visualize the attribution for an output of a network \cite{CAM}. This was later expanded by Grad-CAM, which used the backpropagated gradients to weight the class activation maps \cite{GradCAM}. After Grad-CAM, multiple different methodologies have been introduced to attempt to improve upon Grad-CAM. Notable later CAM methods include, Grad-CAM++ \cite{GradCAMPLUS}, Ablation-CAM \cite{AblationCam}, Layer-CAM \cite{LayerCam}, XGrad-CAM \cite{XGradCam}, and SmoothGrad-CAM++ \cite{SmoothGradCAMPLUS}.

\subsubsection{Evaluation of Explainability Methods}

In 2018, a study into explainability methods was conducted, which determined that most methods were acting as simple edge detectors \cite{SanityChecks}. The experiments showed that produced saliency maps were mostly independent of the model parameters and labels of the training data, except for notably the saliency maps created using Grad-CAM and DeConvNet. The Pointing Game was introduced as a way to measure the class-sensitivity and accuracy of class-guided explainability methods \cite{ExcitationBP}. Finally, ranking methods for pixel importance have been proposed via removal based methods \cite{LRP,Xrai,RISE}.

\section{Method}

We aim to generate methods for assessing how accurate and stable explanations are. For accuracy, the method should be class-specific and concentrated within the segmentation map of an object of the correct class. Stability should be quantified by the consistency of explanations when minor transformations are applied to an image.

\subsection{The Weighting Game}

\begin{figure}[!t]
    \centering
    \includegraphics[width=\textwidth]{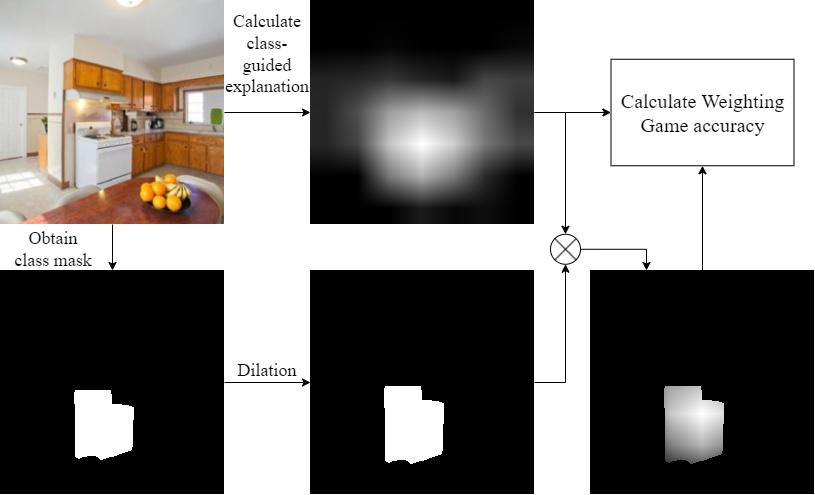}
    \caption{Process of Weighting Game starting from an input image and a class. Firstly, we obtain the class mask for the image and calculate an explanation for the corresponding class. Then, we dilate the class segmentation mask with the produced explanation, which creates a masked explanation. Finally, the mass of the masked explanation is divided with the mass of the explanation, to obtain the accuracy of an explanation.}
    \label{fig:weighting-game-diagram}
\end{figure}

We defined the Weighting Game as a metric of class-guided explainability method accuracy. The prerequisite for Weighting Game is a dataset, which contains class segmentation masks for the classes that the given model is trained to classify. Thus, object detection datasets, such as Pascal Visual Object Classes 2007 \cite{PASCALVOC07} or MS COCO \cite{COCO}, function well for this task.

For calculating one instance of the Weighting Game, we begin with an image and a corresponding class segmentation mask for a class contained within the image. The class segmentation mask is the union of all corresponding class' object segmentation masks found within the image. After obtaining the class segmentation mask, it is dilated using a simple 2D convolution defined with the following equation:
\begin{equation}
\label{eqn:dilation}
    D = M \circledast K,
\end{equation}
where $M$ is the binary class segmentation mask to be dilated, $K$ is a $9\times9$ convolution kernel, and $D$ is the resulting mask. As the resulting mask $D$ is binary, all parts of the mask $M$ that overlap with the convolutional kernel $K$, will be included in the dilated mask $D$. The dilation provides the method with more robustness, as outer edges of the desired objects may also be relevant to the classification of the objects.

Then, using the class corresponding to the segmentation mask, the class-guided explanation is calculated for the image. The resulting explanation is a saliency map, which is either produced in the resolution of the input image or upscaled to it.

The dilated class segmentation mask is then used to measure, how much of the \emph{magnitude} or \emph{mass} of the saliency map is contained within the segmentation mask. This gives a measure of how much of the saliency map is contained within the relevant class' area in the image, thus providing a value for the accuracy of an explainability method. One class/image combination of Weighting Game can be formulated with the following equation:
\begin{equation}
\label{eqn:weighting}
    Accuracy = \frac{\sum_{i,j}S_{i,j}D_{i,j}}{\sum_{i,j}S_{i,j}},
\end{equation}
where $S$ is the class-guided saliency map and $D$ is the corresponding class' dilated segmentation mask. As Equation. \ref{eqn:weighting} provides the accuracy of one class within an image, it will be used to calculate the accuracy of all class/image combinations in the chosen dataset. The final result of Weighting Game is the mean accuracy of all class/image combinations. A diagram of the process of calculating Weighting Game accuracy is visualized in Figure. \ref{fig:weighting-game-diagram}.

\subsection{Explanation Stability}

Stability of an explainability method can be assessed via the consistency of saliency maps when minor transformations are applied to an image. Thus, this factor is important when, e.g. producing class-guided explanations for videos. Stability can characterize the reliability of an explainability method, as when given similar information to an explainability method, it should produce a similar saliency map.

Stability of an explanation method will be assessed via two separate metrics. Firstly, we simulate camera motion by creating 3D effect zoom and pan videos using 3D Ken Burns effect \cite{KenBurns}. Artificially created videos are used, as they can provide a diverse set of videos with a consistent level of motion between frames. This also provides the opportunity to stay within the same distribution of data as in training, as the validation split of the training dataset can be used to generate the videos. The created videos will have 150 frames, of which, the first half consists of a zoom in with ratio 1.5, with the second half consisting of a zoom back to the first frame. Using these videos, we calculate explanations for five pairs of consecutive frames in the video.

\begin{figure}[!t]
    \centering
    \includegraphics[width=\textwidth]{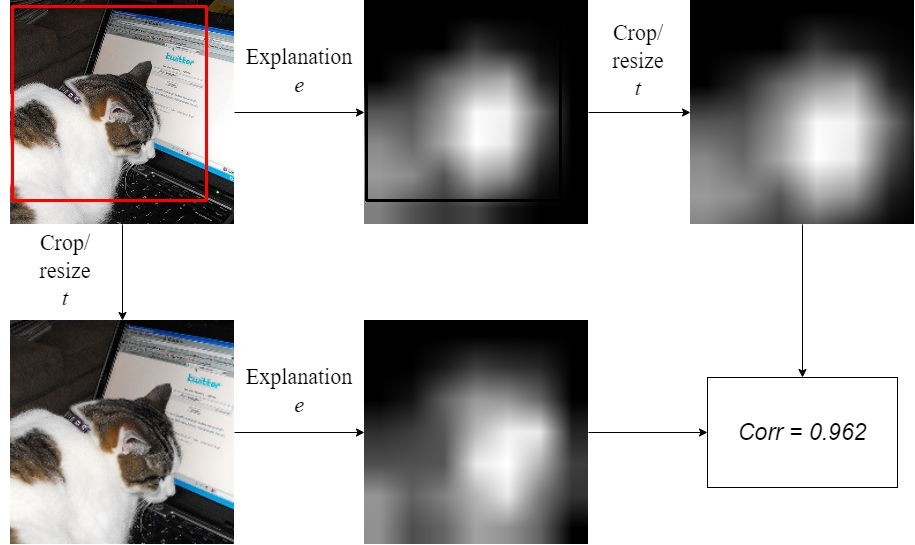}
    \caption{Diagram of explanation stability using a zoom and pan transformation. First, a random resized crop is calculated, which is visualized as red borders in the original image. Then, a class-guided explanation is calculated for the original image, which is then cropped with the same calculated crop. After that, the original image is cropped and a class-guided explanation is calculated for the cropped image. Finally, the correlation between the resulting two saliency maps is calculated. Explanations were processed with class-guided Grad-CAM \cite{GradCAM} and ResNet50 \cite{ResNet} using the class 'cat'.}
    \label{fig:stability-diagram}
\end{figure}

The explanations will be processed using class-guided methods, with the chosen class being the highest probability class for the first frame of the video. Then, the correlation between these saliency maps is calculated, to produce a measure for the stability of an explanation method. If the difference between frames is small enough, where the placement of objects is practically the same, the correlation between the saliency maps should be high. The correlations between saliency maps are calculated using Spearman's rank-order correlation \cite{Spearman}, as in other literature where correlation between saliency maps is evaluated \cite{NormGrad,SaliencyEye}. Thus, the equation used to calculate the stability of a method will be as follows:
\begin{equation}
\label{eqn:stability-video}
    Stability = corr(e(M_i, c), e(M_{i+1}, c)),
\end{equation}
where $M_i$ is the $i$th frame of a video, $c$ is the highest probability class for frame $M_0$, $corr$ is the Spearman rank-order correlation, and $e$ is the class-guided explanation function that takes in an image and a class.

The second method for assessing explanation stability is by calculating a zoom, pan, and upscaling (back to original resolution) transformation on an image. After this, saliency maps are calculated for both images. Then, the saliency map for the original image is transformed in the same manner as the transformed image, as the transformation is greater than those seen with the earlier method using consecutive frames of a video. Finally, the correlations are calculated between the saliency map for the transformed image and the transformed saliency map for the original image. Using the same transformation on the original image and the corresponding saliency map, we can align the saliency maps for the original and transformed images. Therefore, the contents are should be the same, however, with different scale in the image. The stability can be expressed with:
\begin{equation}
\label{eqn:stability-transformation}
    Stability = corr(t(e(M, c)), e(t(M), c)),
\end{equation}
where $t$ is the zoom/pan transformation function, $M$ is the input image, $c$ is the highest probability class for $M$, $corr$ is the Spearman rank-order correlation, and $e$ is the class-guided explanation function. A diagram of the stability metric is shown in Figure. \ref{fig:stability-diagram}.

\section{Experiments}

In this section, we describe the evaluation of various class-guided explanation methods using our accuracy and stability frameworks. Code to reproduce the experiments is available at \url{https://github.com/lassiraa/weighting-game}.

\subsection{Models and Dataset}

The chosen model architectures for experiments are ResNet50 \cite{ResNet}, VGG16-BN \cite{VGG}, ViT-B/32 \cite{ViT}, and Swin-T \cite{Swin}. The models represent some of the most popular architectures from both convolutional and transformer based image classification architectures. The inference times of all models are also comparable, which is why they are desirable to compare to each other. Pre-trained versions of all models were obtained from Torchvision's available models \cite{PyTorch}.

As all the chosen models are trained on the ImageNet \cite{deng2009imagenet} dataset, they need to be fine-tuned in order to be used for the Weighting Game. Thus, the COCO dataset is chosen for fine-tuning the models to a multi-label classification task. The fine-tuning procedure is similar to that of the one implemented in the Pointing Game \cite{ExcitationBP}.

Only the weights of the last layer are adjusted in the fine-tuning procedure, using binary cross entropy loss. For the training procedure, random resized crop and horizontal flips are used for data augmentation. For validation, images are center cropped and resized to $224\times224$ resolution. The Adam optimizer \cite{AdamOpt} is used for optimization, using learning rates of 0.001 and 0.002 for the convolutional and transformer models respectively. Exponential learning rate scheduler was used with 0.95 gamma. All models were trained for 50 epochs, at which point the validation losses had plateaued for all models.

\begin{figure}[!t]
    \centering
    \includegraphics[width=\textwidth]{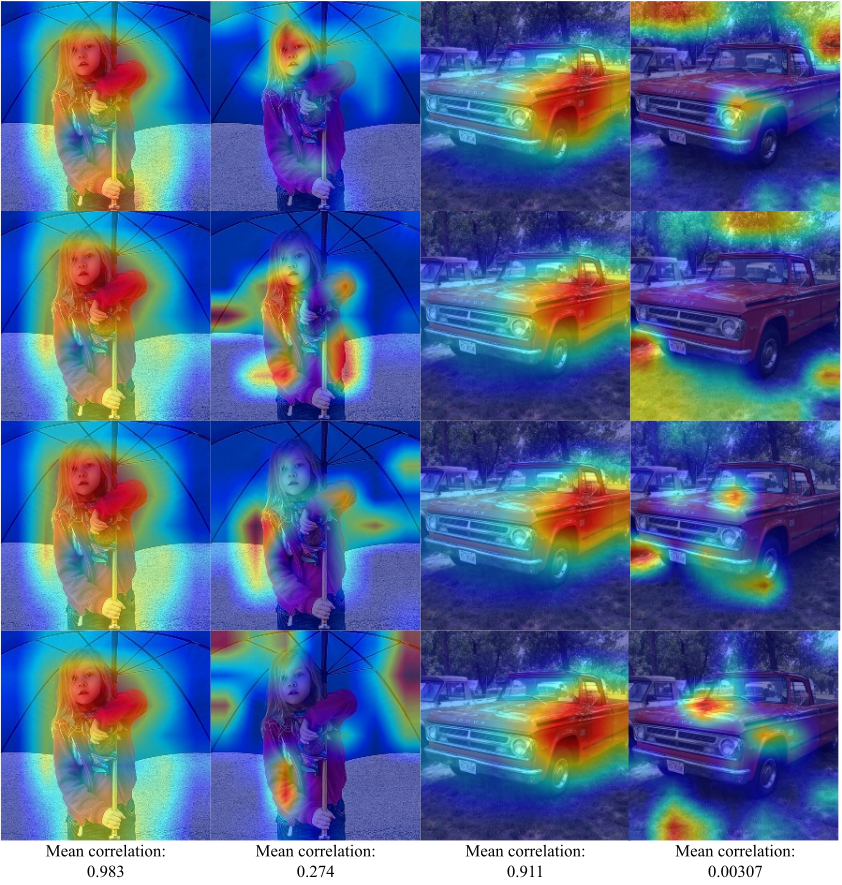}
    \caption{Examples of consecutive frames from created 3D effect videos and their corresponding class-guided explanations. Displayed correlations values are mean correlations between consecutive frame pairs, as a measure of the explainability method's stability. Full videos can be found in the supplementary materials. First and third columns processed with ResNet50 \cite{ResNet}, Grad-CAM \cite{GradCAM}, and class 'person', while second and fourth columns were processed with ViT-B/32 \cite{ViT}, Layer-CAM \cite{LayerCam}, and class 'elephant'}
    \label{fig:video-frames}
\end{figure}

For consistency in models and datasets, the COCO validation set is used in all explanation accuracy and stability evaluations. This way, the distribution of the data remains similar in training and evaluation environments, as we are not attempting to measure the out-of-distribution performance of the models or explanation methods. Examples of produced 3D effect videos can be found in the supplementary materials with saliency maps overlaid on the video, such as in Figure \ref{fig:video-frames}.

\subsection{Saliency Methods}

The chosen saliency methods to assess in this study were Guided Backpropagation \cite{GuidedBackProp}, Grad-CAM \cite{GradCAM}, Grad-CAM++ \cite{GradCAMPLUS}, Layer-CAM \cite{LayerCam}, XGrad-CAM \cite{XGradCam}, and Ablation-CAM \cite{AblationCam}. All methods have implementations available in the Pytorch Grad-CAM package \cite{PytorchGradCam}, which provides consistence in implementations. All methods except Ablation-CAM only require one pass through a model, making them computationally efficient, and interesting to determine whether the added computational cost of Ablation-CAM is worth it. Guided Backpropagation provides a baseline for comparison. All methods also satisfy the requirement of being class-guided.

The target layers for the methods are chosen to be from either the last convolutional or transformer block. For ResNet50, the chosen target layers are the layers of the last identity block, while VGG16-BN has the last max pooling layer used as the target layer. Both the ViT-B/32 and Swin-T models have the last transformer block's first layer normalization as the target layer. All saliency maps are upscaled to the input image's resolution using bilinear interpolation. Transformer models' saliency maps are also reshaped from the flattened patches to the shape of the input image.

\subsection{Results}

\subsubsection{Weighting Game}

\begin{table}[t]
\caption{Explanation accuracy results of different saliency methods for different model architectures. For Weighting Game, the recorded values are the mean percentage of saliency map magnitude contained within the correct class segmentation mask. Weighting Game small is the same metric for objects under 10\% of the size of the image. Results are contrasted to the reproduced Pointing Game \cite{ExcitationBP} metric. Best and second best explainability method results for each model are bolded and underlined, respectively.}
\label{table:weighting-game}
\centering
\begin{tabularx}{\textwidth}{|X|p{0.22\textwidth}|X|X|X|X|}
\hline
 & & \multicolumn{4}{|c|}{Explanation accuracy (\%)} \\
\hline
 & & ResNet50 & VGG16-BN & ViT-B/32 & Swin-T \\
\hline
\multirow{3}{0.12\textwidth}{Ablation-CAM \cite{AblationCam}} & Weighting & 21.1 & 19.3 & 23.7 & \underline{22.2}\\
 & Pointing & 29.3 & 23.2 & 26.3 & \textbf{30.9}\\
 & Weighting small & 4.74 & 4.00 & 5.01 & \underline{4.93} \\
\hline
\multirow{3}{0.12\textwidth}{Grad-CAM \cite{GradCAM}} & Weighting & \textbf{30.5} & \textbf{29.2} & 26.7 & 14.6\\
 & Pointing & \textbf{38.0} & \textbf{36.6} & \underline{32.4} & 15.6\\
 & Weighting small & \underline{11.9} & \textbf{10.4} & 8.81 & 4.17 \\
\hline
\multirow{3}{0.12\textwidth}{Grad-CAM++ \cite{GradCAMPLUS}} & Weighting & 20.4 & 22.0 & \underline{27.7} & 13.2\\
 & Pointing & 29.8 & 33.7 & 32.2 & 13.7\\
 & Weighting small & 5.29 & 6.06 & \underline{9.55} & 4.10\\
\hline
\multirow{3}{0.12\textwidth}{Guided Backprop. \cite{GuidedBackProp}} & Weighting & 24.5 & 21.7 & 16.5 & 16.7\\
 & Pointing & 25.9 & 23.6 & 16.3 & 17.2 \\
 & Weighting small & 8.40 & 7.04 & 3.79 & 4.49 \\
\hline
\multirow{3}{0.12\textwidth}{Layer-CAM \cite{LayerCam}} & Weighting & 20.3 & 22.3 & \textbf{30.6} & \textbf{22.5}\\
 & Pointing & 28.8 & 34.5 & \textbf{34.2} & \underline{30.5}\\
 & Weighting small & 5.20 & 6.17 & \textbf{13.8} & \textbf{10.5} \\
\hline
\multirow{3}{0.12\textwidth}{XGrad-CAM \cite{XGradCam}} & Weighting & \underline{30.5} & \underline{29.0} & 16.2 & 16.1\\
 & Pointing & \textbf{38.0} & \underline{36.4} & 14.6 & 14.3\\
 & Weighting small & \textbf{11.9} & \underline{10.3} & 3.66 & 3.66 \\
\hline
\end{tabularx}
\end{table}

After calculating the results of the Weighting Game, we obtain mean accuracy results displayed in Table. \ref{table:weighting-game}. The table contains results for the Weighting Game using all objects and only objects that are under 10\% of the size of the image. Pointing Game \cite{ExcitationBP} scores are also included for comparison.

\begin{figure}[t]
     \centering
     \begin{subfigure}[t]{0.3\textwidth}
         \centering
         \includegraphics[width=\textwidth]{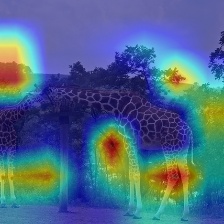}
         \caption{Class-guided explanation with class 'giraffe'.}
     \end{subfigure}
     \hfill
     \begin{subfigure}[t]{0.3\textwidth}
         \centering
         \includegraphics[width=\textwidth]{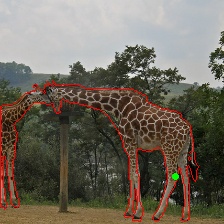}
         \caption{Pointing Game with $acc=100\%$.}
     \end{subfigure}
     \hfill
     \begin{subfigure}[t]{0.3\textwidth}
         \centering
         \includegraphics[width=\textwidth]{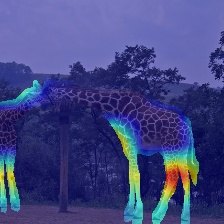}
         \caption{Weighting Game with $acc=19.2\%$.}
     \end{subfigure}
        \caption{Example of Pointing Game's limitations, when saliency map is not concentrated around one central point. Explanation was processed with class-guided XGrad-CAM \cite{XGradCam} and Swin-T \cite{Swin}}
        \label{fig:weighting-game-superior}
\end{figure}

When results of Weighting Game are compared to the ones of Pointing Game, we can immediately see that the values of Pointing Game are higher, implying that the task of Pointing Game is generally easier. When relative values are compared, it appears that generally CNNs tend to perform better with the chosen explainability methods, when using Pointing Game as the metric. This could be due to the very low resolution of CAM methods when using late layers of CNNs. This is emphasized with for example Figure \ref{fig:video-frames}, where saliency maps produced with ResNet50 are much more spread out, when compared to the ones produced by ViT-B/32. Pointing Game fails to quantify this loss in fidelity of saliency maps, as the accuracy values are binary.

Pointing Game is more prone to errors in measuring the accuracy of a saliency map. In cases where the saliency map is mostly concentrated around one point, there is a possibility to narrowly miss the correct segmentation mask, such as in Figure \ref{fig:pointing-game-misses}. Simultaneously, it fails to quantify cases where the saliency map is distributed all over the image, such as in Figure \ref{fig:weighting-game-superior}. In these cases, the classification of an explanation as either a hit or a miss is reductive to the quality of the explanation. Additional Pointing Game and Weighting Game visualizations are provided in the supplementary materials.

From observing the results of the Weighting Game, it shows that model architecture affects the results significantly, with Grad-Cam/XGrad-CAM functioning the best with CNN-based architectures, while Layer-CAM/Grad-CAM++ show great results with ViT-B/32. Swin-T shows poor results with all saliency methods, except for Ablation-CAM and Layer-CAM. Grad-CAM and Grad-CAM++ show results worse than randomly distributed saliency using Swin-T, which shows that the selection of the target layer could be poor or that the saliency methods simply do not function well with the architecture.

While CAM methods were initially designed for CNNs, they do show promise with vision transformers, as ViT-B/32 has the best results with three out of the five CAM methods. The results highlight the need to consider the model architecture in the process of choosing a saliency method, as even within the same family of architectures (CNN/ViT), the results vary significantly.

When only small objects (>10\% of the size of the image) are considered, the results of the Weighting Game change significantly. Notably, the relative accuracy of Layer-CAM with transformer models improves significantly when compared to other methods. Results of Ablation-CAM can also be seen deteriorating significantly for all model architectures. CNNs perform comparatively better in the Pointing Game, which highlights the lack of the full picture of explanation accuracy when using the Pointing Game as a metric. It does not properly capture, how the explanations are distributed, which can lead to results favoring certain method/model combinations over others.

\subsubsection{Explanation Stability Using 3D-Effect Videos}

\begin{table}[t]
\caption{Results of explanation stability. Consecutive frames values are mean correlations between saliency maps of consecutive frames from artificially generated 3D effect videos from still images. Zoom and pan transformation values are mean correlations calculated with Equation. \ref{eqn:stability-transformation}}
\label{table:stability}
\centering
\begin{tabularx}{\textwidth}{|p{0.2\textwidth}|X|X|X|X|X|}
\hline
& & \multicolumn{4}{|c|}{Explanation stability} \\
\hline
& & ResNet50 & VGG16-BN & ViT-B/32 & Swin-T \\
\hline
\multirow{2}{0.2\textwidth}{Grad-CAM \cite{GradCAM}} & 3D-effect & 0.961 & 0.938 & \textbf{0.966} & \textbf{0.878}\\
& Crop & 0.878 & 0.782 & \textbf{0.669} & \textbf{0.593}\\
\hline
\multirow{2}{0.2\textwidth}{Grad-CAM++ \cite{GradCAMPLUS}} & 3D-effect & \underline{0.987} & \textbf{0.984} & \underline{0.943} & 0.747\\
& Crop & \textbf{0.908} & \textbf{0.886} & \underline{0.634} & 0.459\\
\hline
\multirow{2}{0.2\textwidth}{Guided Backprop. \cite{GuidedBackProp}} & 3D-effect & 0.592 & 0.599 & 0.743 & 0.444\\
& Crop & 0.473 & 0.572 & 0.132 & 0.107\\
\hline
\multirow{2}{0.2\textwidth}{Layer-CAM \cite{LayerCam}} & 3D-effect & \textbf{0.987} & \underline{0.983} & 0.928 & \underline{0.845}\\
& Crop & \underline{0.908} & \underline{0.882} & 0.505 & \underline{0.475} \\
\hline
\multirow{2}{0.2\textwidth}{XGrad-CAM \cite{XGradCam}} & 3D-effect & 0.961 & 0.938 & 0.0732 & 0.0718\\
& Crop & 0.878 & 0.777 & 0.0238 & 0.0252\\
\hline
\end{tabularx}
\end{table}

Table. \ref{table:stability} shows results of explanation stability using 3D-effect videos and zoom/pan transformations. Ablation-CAM was omitted from the stability experiments due to the high inference time of the method. The results of the video stability show that Grad-CAM++/Layer-CAM are most stable for CNNs, slightly ahead of Grad-CAM/ XGrad-CAM, which are the most accurate in the results of the Weighting Game. This shows that stability does not necessarily result in greater accuracy. However, high stability does seem like a prerequisite for good quality explanations, as all methods with clearly worse stability show poor results in explanation accuracy. Interestingly, ViT-B/32 also shows worse stability with Layer-CAM when contrasted to the CNN models, while the accuracy of the method is significantly superior with ViT-B/32.

\subsubsection{Explanation Stability Using Zoom and Pan Transformation}

Pan and zoom effect transformations were created using a random resized crop with crop scale 0.75-0.9, 1:1 aspect ratio, and bilinear interpolation. For every image in COCO validation split, a random resized crop is applied, and saliency maps for the highest probability class are processed for both images. Then, the same crop is applied for the source image's saliency map. Finally, the correlation is calculated between the cropped saliency map and the cropped image's saliency maps, as the saliency maps should align at this point. Table. \ref{table:stability} shows the mean correlations for the method/model combinations.

The mean correlations are significantly lower than those found in the consecutive frames approach, which is sensible as the step taken in this method is far greater. However, the ranks of results also change significantly. Notably, the relative stability of transformer models drops clearly. This effect is especially seen with Layer-CAM and Guided Backpropagation, however, it is present with all other methods. The reason for this behavior could be explained via how the receptive fields operate within transformer based models. This is due to transformer receptive fields being more concentrated locally, when contrasting ViT-B/32 and ResNet50 receptive fields at later layers \cite{DoViTsSeeLikeCNNs}. More concentrated receptive field could yield to results that are less stable, but perhaps more capable of correctly explaining smaller objects, as seen in results of the Weighting Game with small objects in Table. \ref{table:weighting-game}.

\section{Conclusion}

We introduced a new method for more comprehensively assessing the accuracy of a class-guided saliency method, which calculates how much of the weight is contained within the correct class' segmentation mask. We also introduced two ways of assessing an explanation method's stability via transforming an image, either by a sufficiently small transformation or by a slightly larger zoom and pan operation. Our experiments show that, generally, sufficient stability is a prerequisite for accurate saliency maps. However, better stability does not always yield better results in explanation accuracy, as especially small objects are not always detected well with highly stable methods. We also highlight the importance of considering model architecture when choosing a saliency method, as the choice of model architecture can significantly affect which saliency methods give the best results.

\section{Acknowledgement}

%
%
%
\bibliographystyle{splncs04}
\bibliography{mybibliography}

\begin{thebibliography}{10}
\providecommand{\url}[1]{\texttt{#1}}
\providecommand{\urlprefix}{URL }
\providecommand{\doi}[1]{https://doi.org/#1}

\bibitem{SanityChecks}
Adebayo, J., Gilmer, J., Muelly, M., Goodfellow, I., Hardt, M., Kim, B.: Sanity
  checks for saliency maps. In: Proceedings of the 32nd International
  Conference on Neural Information Processing Systems. p. 9525–9536. NIPS'18,
  Curran Associates Inc., Red Hook, NY, USA (2018)

\bibitem{LRP}
Bach, S., Binder, A., Montavon, G., Klauschen, F., M{\"u}ller, K.R., Samek, W.:
  On pixel-wise explanations for non-linear classifier decisions by layer-wise
  relevance propagation. PLoS ONE  \textbf{10} (2015)

\bibitem{GradCAMPLUS}
Chattopadhay, A., Sarkar, A., Howlader, P., Balasubramanian, V.N.: Grad-cam++:
  Generalized gradient-based visual explanations for deep convolutional
  networks. In: 2018 IEEE Winter Conference on Applications of Computer Vision
  (WACV). pp. 839--847 (2018). \doi{10.1109/WACV.2018.00097}

\bibitem{deng2009imagenet}
Deng, J., Dong, W., Socher, R., Li, L.J., Li, K., Fei-Fei, L.: Imagenet: A
  large-scale hierarchical image database. In: 2009 IEEE conference on computer
  vision and pattern recognition. pp. 248--255. Ieee (2009)

\bibitem{AblationCam}
Desai, S., Ramaswamy, H.G.: Ablation-cam: Visual explanations for deep
  convolutional network via gradient-free localization. In: 2020 IEEE Winter
  Conference on Applications of Computer Vision (WACV). pp. 972--980 (2020).
  \doi{10.1109/WACV45572.2020.9093360}

\bibitem{ViT}
Dosovitskiy, A., Beyer, L., Kolesnikov, A., Weissenborn, D., Zhai, X.,
  Unterthiner, T., Dehghani, M., Minderer, M., Heigold, G., Gelly, S.,
  Uszkoreit, J., Houlsby, N.: An image is worth 16x16 words: Transformers for
  image recognition at scale. In: International Conference on Learning
  Representations (2021)

\bibitem{ExpectedGradients}
Erion, G., Janizek, J.D., Sturmfels, P., Lundberg, S.M., Lee, S.I.: Improving
  performance of deep learning models with axiomatic attribution priors and
  expected gradients. Nature Machine Intelligence  \textbf{3}(7),  620--631
  (May 2021). \doi{10.1038/s42256-021-00343-w},
  \url{https://doi.org/10.1038/s42256-021-00343-w}

\bibitem{PASCALVOC07}
Everingham, M., Van~Gool, L., Williams, C.K.I., Winn, J., Zisserman, A.: The
  {PASCAL} {V}isual {O}bject {C}lasses {C}hallenge 2007 {(VOC2007)} {R}esults.
  http://www.pascal-network.org/challenges/VOC/voc2007/workshop/index.html

\bibitem{XGradCam}
Fu, R., Hu, Q., Dong, X., Guo, Y., Gao, Y., Li, B.: Axiom-based grad-cam:
  Towards accurate visualization and explanation of cnns (2020).
  \doi{10.48550/ARXIV.2008.02312}, \url{https://arxiv.org/abs/2008.02312}

\bibitem{PytorchGradCam}
Gildenblat, J., contributors: Pytorch library for cam methods.
  \url{https://github.com/jacobgil/pytorch-grad-cam} (2021)

\bibitem{ResNet}
He, K., Zhang, X., Ren, S., Sun, J.: Deep residual learning for image
  recognition. In: Proceedings of the IEEE conference on computer vision and
  pattern recognition. pp. 770--778 (2016)

\bibitem{LayerCam}
Jiang, P.T., Zhang, C.B., Hou, Q., Cheng, M.M., Wei, Y.: Layercam: Exploring
  hierarchical class activation maps for localization. IEEE Transactions on
  Image Processing  \textbf{30},  5875--5888 (2021).
  \doi{10.1109/TIP.2021.3089943}

\bibitem{Xrai}
Kapishnikov, A., Bolukbasi, T., Vi{\'e}gas, F., Terry, M.: Xrai: Better
  attributions through regions. In: Proceedings of the IEEE/CVF International
  Conference on Computer Vision. pp. 4948--4957 (2019)

\bibitem{AdamOpt}
Kingma, D.P., Ba, J.: Adam: {A} method for stochastic optimization. In: Bengio,
  Y., LeCun, Y. (eds.) 3rd International Conference on Learning
  Representations, {ICLR} 2015, San Diego, CA, USA, May 7-9, 2015, Conference
  Track Proceedings (2015), \url{http://arxiv.org/abs/1412.6980}

\bibitem{COCO}
Lin, T.Y., Maire, M., Belongie, S., Hays, J., Perona, P., Ramanan, D.,
  Doll{\'a}r, P., Zitnick, C.L.: Microsoft coco: Common objects in context. In:
  Fleet, D., Pajdla, T., Schiele, B., Tuytelaars, T. (eds.) Computer Vision --
  ECCV 2014. pp. 740--755. Springer International Publishing, Cham (2014)

\bibitem{Swin}
Liu, Z., Ning, J., Cao, Y., Wei, Y., Zhang, Z., Lin, S., Hu, H.: Video swin
  transformer. In: Proceedings of the IEEE/CVF Conference on Computer Vision
  and Pattern Recognition. pp. 3202--3211 (2022)

\bibitem{Spearman}
Myers, L., Sirois, M.J.: Spearman correlation coefficients, differences
  between. Encyclopedia of statistical sciences  \textbf{12} (2004)

\bibitem{KenBurns}
Niklaus, S., Mai, L., Yang, J., Liu, F.: 3d ken burns effect from a single
  image. ACM Transactions on Graphics  \textbf{38}(6),  184:1--184:15 (2019)

\bibitem{SmoothGradCAMPLUS}
Omeiza, D., Speakman, S., Cintas, C., Weldemariam, K.: Smooth grad-cam++: An
  enhanced inference level visualization technique for deep convolutional
  neural network models. ArXiv  \textbf{abs/1908.01224} (2019)

\bibitem{PyTorch}
Paszke, A., Gross, S., Massa, F., Lerer, A., Bradbury, J., Chanan, G., Killeen,
  T., Lin, Z., Gimelshein, N., Antiga, L., Desmaison, A., Kopf, A., Yang, E.,
  DeVito, Z., Raison, M., Tejani, A., Chilamkurthy, S., Steiner, B., Fang, L.,
  Bai, J., Chintala, S.: Pytorch: An imperative style, high-performance deep
  learning library. In: Wallach, H., Larochelle, H., Beygelzimer, A.,
  d\textquotesingle Alch\'{e}-Buc, F., Fox, E., Garnett, R. (eds.) Advances in
  Neural Information Processing Systems 32, pp. 8024--8035. Curran Associates,
  Inc. (2019),
  \url{http://papers.neurips.cc/paper/9015-pytorch-an-imperative-style-high-performance-deep-learning-library.pdf}

\bibitem{RISE}
Petsiuk, V., Das, A., Saenko, K.: Rise: Randomized input sampling for
  explanation of black-box models. In: BMVC (2018)

\bibitem{DoViTsSeeLikeCNNs}
Raghu, M., Unterthiner, T., Kornblith, S., Zhang, C., Dosovitskiy, A.: Do
  vision transformers see like convolutional neural networks? Advances in
  Neural Information Processing Systems  \textbf{34},  12116--12128 (2021)

\bibitem{NormGrad}
Rebuffi, S.A., Fong, R., Ji, X., Vedaldi, A.: There and back again: Revisiting
  backpropagation saliency methods. In: Proceedings of the IEEE/CVF Conference
  on Computer Vision and Pattern Recognition. pp. 8839--8848 (2020)

\bibitem{GradCAM}
Selvaraju, R.R., Cogswell, M., Das, A., Vedantam, R., Parikh, D., Batra, D.:
  Grad-cam: Visual explanations from deep networks via gradient-based
  localization. In: Proceedings of the IEEE international conference on
  computer vision. pp. 618--626 (2017)

\bibitem{SimonyanSaliency}
Simonyan, K., Vedaldi, A., Zisserman, A.: Deep inside convolutional networks:
  Visualising image classification models and saliency maps. CoRR
  \textbf{abs/1312.6034} (2014)

\bibitem{VGG}
Simonyan, K., Zisserman, A.: Very deep convolutional networks for large-scale
  image recognition. In: Bengio, Y., LeCun, Y. (eds.) 3rd International
  Conference on Learning Representations, {ICLR} 2015, San Diego, CA, USA, May
  7-9, 2015, Conference Track Proceedings (2015),
  \url{http://arxiv.org/abs/1409.1556}

\bibitem{SmoothGrad}
Smilkov, D., Thorat, N., Kim, B., Viégas, F., Wattenberg, M.: Smoothgrad:
  removing noise by adding noise (2017). \doi{10.48550/ARXIV.1706.03825},
  \url{https://arxiv.org/abs/1706.03825}

\bibitem{GuidedBackProp}
Springenberg, J.T., Dosovitskiy, A., Brox, T., Riedmiller, M.A.: Striving for
  simplicity: The all convolutional net. CoRR  \textbf{abs/1412.6806} (2015)

\bibitem{IntegratedGradients}
Sundararajan, M., Taly, A., Yan, Q.: Axiomatic attribution for deep networks.
  In: Proceedings of the 34th International Conference on Machine Learning -
  Volume 70. p. 3319–3328. ICML'17, JMLR.org (2017)

\bibitem{SaliencyEye}
Volokitin, A., Gygli, M., Boix, X.: Predicting when saliency maps are accurate
  and eye fixations consistent. In: Proceedings of the ieee conference on
  computer vision and pattern recognition. pp. 544--552 (2016)

\bibitem{DeConvNet}
Zeiler, M.D., Fergus, R.: Visualizing and understanding convolutional networks.
  In: ECCV (2014)

\bibitem{ExcitationBP}
Zhang, J., Bargal, S.A., Lin, Z., Brandt, J., Shen, X., Sclaroff, S.: Top-down
  neural attention by excitation backprop. International Journal of Computer
  Vision  \textbf{126}(10),  1084--1102 (Dec 2017).
  \doi{10.1007/s11263-017-1059-x},
  \url{https://doi.org/10.1007/s11263-017-1059-x}

\bibitem{CAM}
Zhou, B., Khosla, A., Lapedriza, A., Oliva, A., Torralba, A.: Learning deep
  features for discriminative localization. In: Proceedings of the IEEE
  conference on computer vision and pattern recognition. pp. 2921--2929 (2016)

\end{thebibliography}

\end{document}